\documentclass[letterpaper, 10 pt, conference]{ieeeconf}
\IEEEoverridecommandlockouts                              
\usepackage[utf8]{inputenc} 
\usepackage[T1]{fontenc}    
\usepackage{url}            
\usepackage{booktabs}       
\usepackage{amsfonts}       
\usepackage{nicefrac}       
\usepackage{microtype}      
\usepackage{xcolor}         
\usepackage{graphicx,epstopdf}

\usepackage{times}
\usepackage{multicol}
\usepackage[bookmarks=true]{hyperref}
\usepackage{bm}
\usepackage{amssymb,amsthm}
\usepackage{amsmath}
\usepackage{mathrsfs}
\usepackage{tabularx} 
\usepackage[linesnumbered,boxed,ruled,commentsnumbered]{algorithm2e}

\usepackage{acronym}
\usepackage{todonotes}
\usepackage{threeparttable}
\usepackage{color}
\usepackage{mathtools}
\usepackage{multirow}
\usepackage[caption=false,font=footnotesize]{subfig}
\usepackage{xcolor} 
\usepackage{soul}
\usepackage{authblk}

\newif\ifuseboldmathops
\newif\ifuseittextabbrevs
\useboldmathopstrue   

\ifuseittextabbrevs
	
	\newcommand{\eg}{{\it e.g. }}
	\newcommand{\ie}{{\it i.e. }}

\else
	
	\newcommand{\eg}{e.g. }
	\newcommand{\ie}{i.e. }

\fi

\ifuseboldmathops

\else

\fi

\ifuseboldmathops

\else

\fi

\ifuseboldmathops

\else

\fi

\ifuseboldmathops

\else

\fi

\newcommand{\union}{\bigcup}

\newcommand{\calL}{\mathcal{L}}
\newcommand{\calF}{\mathcal{F}}
\newcommand{\calN}{\mathcal{N}}
\newcommand{\calT}{\mathcal{T}}

\newcommand{\calA}{\mathcal{A}}
\newcommand{\calS}{\mathcal{S}}
\newcommand{\calG}{\mathcal{G}}
\newcommand{\calE}{\mathcal{E}}

\newcommand{\calH}{\mathcal{H}}
\newcommand{\calB}{\mathcal{B}}

\newcommand{\calV}{\mathcal{V}}

\newcommand{\iactor}[1]{\pi^{\boldtheta_{#1}}}

\newcommand{\ivalue}[1]{V^{\boldphi_{#1}}}
\newcommand{\pstate}[2]{\tilde{s}^{#1}_{#2}}

\newcommand{\grad}{\nabla}
\newcommand{\boldtheta}{\boldsymbol{\theta}}

\newcommand{\boldlambda}{\boldsymbol{\lambda}}
\newcommand{\boldphi}{\boldsymbol{\phi}}
\newcommand{\boldomega}{\boldsymbol{\omega}}

\renewcommand{\vec}[1]{\bm{#1}}

\newcommand{\pts}[1]{\tilde{\vec{#1}}_\calT}

\newcommand{\safe}{\text{safe}}

\newcommand{\KLS}{\text{KEEP-LANE-MAX}}
\newcommand{\CLL}{\text{CHANGE-LANE-LEFT}}
\newcommand{\CLR}{\text{CHANGE-LANE-RIGHT}}

\newcommand{\wq}{\underline{Q}}
\newcommand{\iwq}[1]{\underline{Q}^{\boldomega_{#1}}}
\newcommand{\SafeS}{\textit{Safety Shield}}

\acrodef{mdp}[MDP]{Markov decision process}
\acrodef{dfa}[DFA]{deterministic finite-state automaton}
\acrodef{ltl}[LTL]{linear temporal logic}
\acrodef{ltlf}[LTL$(\calF)$]{quantitative linear temporal logic}
\acrodef{ag}[AG]{Assume-Guarantee}
\acrodef{ssp}[SSP]{Stochastic Shortest Path}
\acrodef{mcmc}[mcmc]{Monte Carlo Markov chain}

\theoremstyle{definition}
 \newtheorem{definition}{Definition}
 \newtheorem{example}{Example}

\acrodef{ltl}[LTL]{linear temporal logic formula}
\acrodef{mdp}[MDP]{Markov decision process}
\acrodef{smdp}[Semi-MDP]{Semi-Markov decision process}

\acrodef{scltl}[scLTL]{syntactically co-safe LTL}

\newcommand\zhili[1]{\textcolor{purple}{#1}}

\usepackage{soul}

\title{Safety Guaranteed Robust Multi-Agent Reinforcement Learning with Hierarchical Control for Connected and Automated Vehicles}

\author{Zhili Zhang$^{*1}$\quad  H M Sabbir Ahmad$^{*2}$\quad  Ehsan Sabouni$^{*2}$\quad  Yanchao Sun$^{3}$ \\
Furong Huang$^{3}$  \quad Wenchao Li$^{2}$\quad Fei Miao$^{1}$
\thanks{*Z.~Zhang, H. M. S.~Ahmad and E.~Sabouni contributed equally}
\thanks{$^{1}$Z.~Zhang, and F.~Miao are with the Department of Computer Science and Engineering, University of Connecticut, Storrs Mansfield, CT, USA 06268 {\tt\small \{zhili.zhang, fei.miao\}@uconn.edu}}
\thanks{$^{2}$H. M. S.~Ahmad, E.~Sabouni, and W.~Li are with the Division of Systems Engineering and Department of Electrical \& Computer Engineering, Boston University, Boston, MA, USA 02215{\tt\small \{sabbir92, esabouni, wenchao\}@bu.edu}}
\thanks{$^{3}$Y.~Sun, and F.~Huang are with the Department of Computer Science, University of Maryland, College Park, MD, USA 20742 {\tt\small \{ycs, furongh\}@umd.edu.} } 
}

\begin{document}


\maketitle


\begin{abstract}
We address the problem of coordination and control of Connected and Automated Vehicles (CAVs) in the presence of imperfect observations in mixed traffic environment. A commonly used approach is learning-based decision-making, such as reinforcement learning (RL). However, most existing safe RL methods suffer from two limitations: (i) they assume accurate state information, and (ii) safety is generally defined over the expectation of the trajectories. It remains challenging to design optimal coordination between multi-agents while ensuring hard safety constraints under system state uncertainties (e.g., those that arise from noisy sensor measurements, communication, or state estimation methods) at every time step. We propose a safety guaranteed hierarchical coordination and control scheme called Safe-RMM to address the challenge. Specifically, the high-level coordination policy of CAVs in mixed traffic environment is trained by the Robust Multi-Agent Proximal Policy Optimization (RMAPPO) method. Though trained without uncertainty, our method leverages a worst-case Q network to ensure the model's robust performances when state uncertainties are present during testing. The low-level controller is implemented using model predictive control (MPC) with robust Control Barrier Functions (CBFs) to guarantee safety through their forward invariance property. We compare our method with baselines in different road networks in the CARLA simulator. Results show that our method provides best evaluated safety and efficiency in challenging mixed traffic environments with uncertainties.
\end{abstract}



\section{Introduction}
\label{sec:intro}
\begin{figure}[!t]
\centering
\scalebox{.98}
{
    \subfloat[]
    {\centering\label{fig:cross_safe}\includegraphics[height=3.2cm,width=3.2cm]{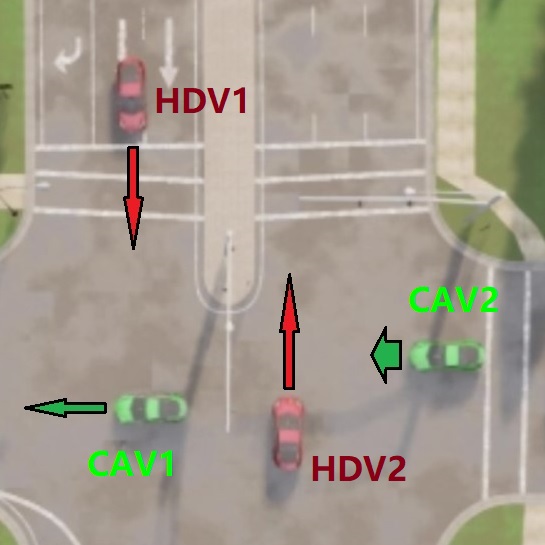}} 
}
\qquad
\scalebox{.98}
{
    \subfloat[]
    {\centering\label{fig:cross_collision}\includegraphics[height=3.2cm,width=3.2cm]{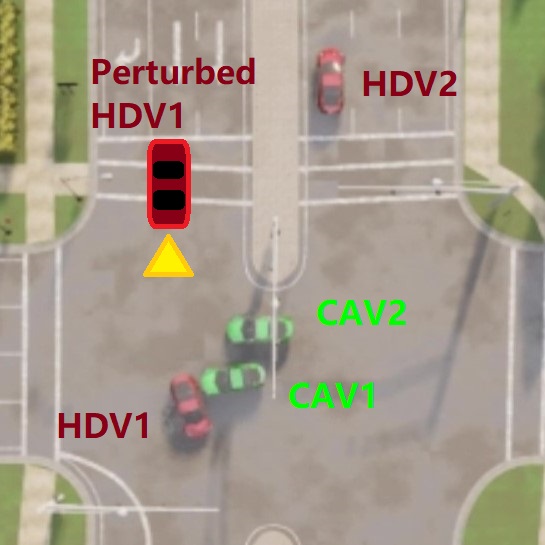}}
}
\vspace*{-0.2cm}
\caption{\textit{Intersection} (a, b): two HDVs run the red light when two CAVs are passing the box in \textit{Intersection} scenario. CAVs are in green; HDVs are in red. (\ref{fig:cross_safe}): CAVs successfully avoid collision with our method during testing with state uncertainties; (\ref{fig:cross_collision}): CAV1 adopting benchmark (MCP) collides with an HDV because the perturbed location of HDV1 (with yellow triangle) misleads the CAV to believe collision will not happen.}
\label{fig:Scenarios}
\vspace*{-8pt}
\end{figure}

Machine learning models assisted by more accurate on-board sensors, such as camera and LiDARs, have enabled intelligent driving to a certain degree. Meanwhile, advances in wireless communication technologies also make it possible for information sharing beyond an individual's perception~\cite{martin2020low, mun2021secure}. Through vehicle-to-everything (V2X) communications, it has been shown that shared information can contribute to CAVs decision-making~\cite {buckman2020generating,miller2020cooperative, han2022stable}, and improve the safety and coordination of CAVs~\cite{Coordinate_CAV, CV_intersection, zhang2023spatial}.  

 
However, it remains challenging for reinforcement learning (RL) or multi-agent reinforcement learning (MARL)-based decision-making methods to guarantee the safety of CAVs in complicated dynamic environments containing human driven vehicles (HDVs) and optimize the joint behavior of the entire system. For real-world CAVs, state uncertainties that may result from noisy sensor measurements, state estimation algorithms or the communication medium pose another challenge. There can be scenarios where safety is highly correlated with the correctness of state information especially in the presence of HDVs/unconnected vehicles, as shown in Fig. ~\ref{fig:cross_collision}.

In this work, we propose a hierarchical decision-making approach for the coordination and control problem of CAVs in mixed traffic environments. At the top of the hierarchy is a Robust MARL policy that learns the cooperative behavior of CAVs by generating discrete planning actions for each vehicle. At the lower level, to guarantee the safety of each CAV, a MPC controller with control barrier function (CBF) constraints is designed to track the planned path according to the MARL action. Specifically, for cooperative policy-learning, we design a robust MAPPO (RMAPPO) algorithm that optimizes the worst-case Q network~\cite{liang2022efficient} as a critic allowing the MAPPO~\cite{yu2022surprising} to train a robust policy without simulating potential state uncertainties during the training process. The MPC controller using robust CBFs serves twofold purposes: guarantees safety in the presence of state uncertainties for the CAVs in mixed traffic environment; and tracks the planned path determined by the RMAPPO policy's actions. 
In summary, the main contributions of this work are:


\begin{itemize}

\item We propose a hierarchical decision-making framework, Safe-RMM, for CAVs in mixed traffic environments. The framework comprises of two levels whereby the top level is Robust MARL (the "RM" in Safe-RMM) that determines discrete actions conditioned on the behavior of other CAVs and HDVs. The low-level controller uses MPC (the final "M" in Safe-RMM) with CBFs to execute the high-level plan while guaranteeing safety to the neighborhood vehicles through the forward invariance property of CBFs. 


\item To handle states uncertainties, we design the robust MARL algorithm which only requires training one more critic for the agents but no prior knowledge of uncertainties. Additionally, the MPC controller is incorporated with robust CBFs to consistently generate safe controls given MARL decisions, and to be endowed with the robustness against erroneous system state. 


\item We validate through experiments in CARLA simulator that the proposed Safe-RMM approach significantly improves the collision-free rate and allows the CAV agents to achieve higher overall returns compared to baseline methods. Ablation studies further highlight the contributions of both the robust MARL algorithm and the MPC-CBF controller, as well as their reciprocal effects. 
\end{itemize}

\section{Related Work}
\label{sec:related}

In this section we provide a survey of the existing literature in this area along with their limitations to motivate our proposed approach.

\paragraph{Safe RL and Robust RL} Different approaches have been proposed to guarantee or improve safety of the system, such as defining a safety shield or barrier assisting RL or MARL algorithm in either training or execution stage~\cite{brunke2022safe,MARLShield_21,cai2021safe}, constrained RL/MARL that learns a risk network~\cite{wen2020safe}, an expected cost function~\cite{lu2021decentralized, safeMARL_AI23}, or cost constraints from language~\cite{LLMsafe_24} that define the safety requirements. For MARL of CAV, safety-checking module with CBF-PID controller for each individual vehicle has been designed~\cite{zhang2023spatial, wang2023multi,han2022behavior}. However, the above works assume \textit{accurate state inputs} to RL or MARL algorithm from the driving environment and cannot tolerate noisy or inaccurate state input. 
Meanwhile, \textit{robust RL and robust MARL} that only considers to train a policy under state uncertainty or model uncertainty ~\cite{liang2022efficient,han2022solution, he2023robust, salvato2021crossing, pmlr-v70-pinto17a} without explicitly considering the safety requirements have been proposed recently.  However, in the multi-agent settings with imperfect observations, considering both safety requirements and robustness in an unified decision-making framework for CAVs still remains challenging.


\paragraph{Rule-Based Approaches} Unified optimization framework poses challenges that can be addressed by decomposing the problem into hierarchical structures. Specifically, the higher level control is responsible for decision making and the lower level control is responsible for safe execution. For the higher level planner, heuristic rule based methods can be employed in which a set of rules govern the behavior of each agent within the system. For instance existing driving behavior models in mixed traffic can be found in \cite{Treiber2000}, \cite{Ksting2007},\cite{MUNIGETY2018284,Olstam2004ComparisonOC}. However, these models often lack robustness and make various assumptions about HDVs, which prevents generalization to all scenarios. 
MPC can be used for the lower level controller due to its ability in reference tracking and handling hard constraints in real time. In situations where imperfect observations are present, robust MPC approaches may be used, such as tube MPC ~\cite{lopez2019dynamic,MAYNE200736}. Nevertheless, tube-based MPC approaches require a feedback controller that can keep the actual system trajectory close to the nominal one. The calculation of such feedback controller is not trivial in multi-agent systems with nonlinear dynamics. Min-Max MPC ~\cite{RAIMONDO20095} can also be adopted but it is often difficult to solve, and when it is approximated, the approximation can result in an overly conservative solution.

In this work, we consider safe and robust coordination and control for multi-agent CAV systems in mixed traffic-environments with state uncertainties.  We define safety as collision-free condition for CAVs, and the concept of robustness refers to agents' capability of ensuring its performances, including safety and efficiency, with state uncertainties. The proposed hierarchical scheme involves robust MARL that works in tandem with a low-level MPC controller using robust CBFs to guarantee safety for CAVs under input state uncertainties. Our robust MARL algorithm does not require injecting perturbations during training.

\section{Problem Formulation}
\label{sec:prob}
\subsection{Problem Description}
\label{sec:prob_desc}
We consider the robust cooperative policy-learning problem under uncertain state inputs for CAVs in mixed traffic environments including HDVs that do not communicate or coordinate with CAVs, and various driving scenarios such as multi-lane intersection and highway (as shown in Fig. \ref{fig:Scenarios}\&\ref{fig:intersection_cases}). We assume that each CAV can get shared information from V2V and V2I communications. We consider that a CAV agent $i$ has accurate self-observation of its driving state but potentially perturbed observations of the other vehicles. The two parts collectively constitute its state $s_i$ in reinforcement learning explained in Sec.~\ref{sec:SPMARL_formulation}, and also used by the MPC controller as inputs to the robust CBFs.

\subsection{Formulation of MARL with State Uncertainty for CAVs}
\label{sec:SPMARL_formulation}
The problem of Multi-Agent Reinforcement Learning with State Uncertainty for CAVs is defined as a tuple $G = (\calS, \calA, P, \{r_i\}, \tilde{o}, \calG, \gamma)$ where $\calG\coloneqq(\calN, \calE)$ is the communication network of all CAV agents. 
$\calS$ is the joint state space of all agents: $\calS:=\calS_1 \times \dots \times \calS_n$. The state space of agent $i$: $\calS_i = \{o_i, o_{\calN_i}, o_{\calN_i^{UV}}\}$ contains self-observation $o_i$, observations $o_{\calN_i}= \{o_j|j\in \calN_i\}$ being the communicated message shared by neighbor connected agents $\calN_i$, observations $o_{\calN_i^{UV}}$ of unconnected vehicles $\calN_i^{UV}$ either observed by agent $i$ itself or shared by other agents or infrastructures. For example, self-observation $o_i$ and shared observations $o_{\calN_i}$ can contain location, velocity, acceleration and lane-detection: $(\vec{l}, \vec{v}, \vec{\alpha}, LD)$; observations of unconnected vehicles $\calN_i^{UV}$ can contain location and velocity $(\vec{l}, \vec{v})$. 

In this work, we consider agent $i$ suffers from uncertain observed locations and velocities of other vehicles aside from the ego vehicle (\ie $\tilde{s}_i = \{o_i\} \union \{\tilde{o}_j | j \in \calN_i \union \calN_i^{UV}\}$), where $\tilde{o}_j$ denotes an uncertain observation over vehicle $j$ in comparison with the true accurate self-observation $o_i$. The uncertainty is defined by bounded errors $(e_l, e_v)\coloneqq \tilde{o} = (\tilde{\vec{l}}, \tilde{\vec{v}}), \tilde{\vec{l}}=\vec{l} + e_l, \tilde{\vec{v}}=\vec{v}+ e_v$. The implementation of state uncertainty in testing experiments is explained in Sec.~\ref{sec:experiment}. 


The joint action set is $\calA:= \calA_1\times \cdots \times \calA_n$ where $\calA_i = \{a_{i,1}, a_{i,2}, \cdots, a_{i,3+k} \}$ is the discrete finite action space for agent $i$. $a_{i,1}$: $\KLS$ - the CAV $i$ maximize its reference throttle in the current lane. $a_{i,2}$: $\CLL$ - the CAV $i$ changes to its left lane. $a_{i,3}$: $\CLR$ - the CAV $i$ changes to its right lane. In experiment, the path planner will set a target waypoint trajectory onto its left/right neighboring lane. $a_{i,4}, \ldots, a_{i,3+k}$ are $k$ lane-keeping actions associated with different reference throttle values. By choosing action $a_{i,5}$, for example, reference throttle value $throttle_5$ will first converted into reference acceleration then it will be fed to the controller and the calculation of safe control is introduced in~\ref{sec:Robust_MPC}. The state transition function is $P: \calS\times \calA \times \calS \mapsto [0,1]$.
The reward functions $\{r_i\coloneqq \calS \times \calA \mapsto\mathbb{R} \}$ are defined as

\footnotesize
\begin{align}
    r_i(s,a) =&\sum_{j}\mu^{v}_{i,j}\|\vec{v}_j\|_2 + \sum_{j}\mu^{l}_{i,j}\|\vec{l}_j - \vec{d}_j\|_2 +
    \sum_{j}\mu^{s}_{i,j}r^s_j(s,a)
    \label{reward_func}
\end{align}
\normalsize
in which $\vec{v}_j$ is vehicle $j$'s velocity; $\vec{d}_j$ is $j$'s default destination and $r^s_j$ is the safety reward. $\mu^{v}, \mu^{l}, \mu^{s}$ are non-negative weights balancing the proportions of individual and total achievement. As break-downs of the safety reward $r^s_j(s,a)=p^{\textit{Col}}(s,a)+p^{\textit{MPC}}(s,s_{\calN_i},a)$, the collision penalty $p^{\textit{Col}}(s,a)$ penalizes collision and $p^{\textit{MPC}}(s,s_{\calN_i},a)$ penalizes the infeasiblity of the low level controller.

\section{Methodology}\label{sec:Methodology}
\begin{figure*}[!t]
    \centering
    \scalebox{0.85}[0.85]
    {\includegraphics[width=0.95\linewidth]{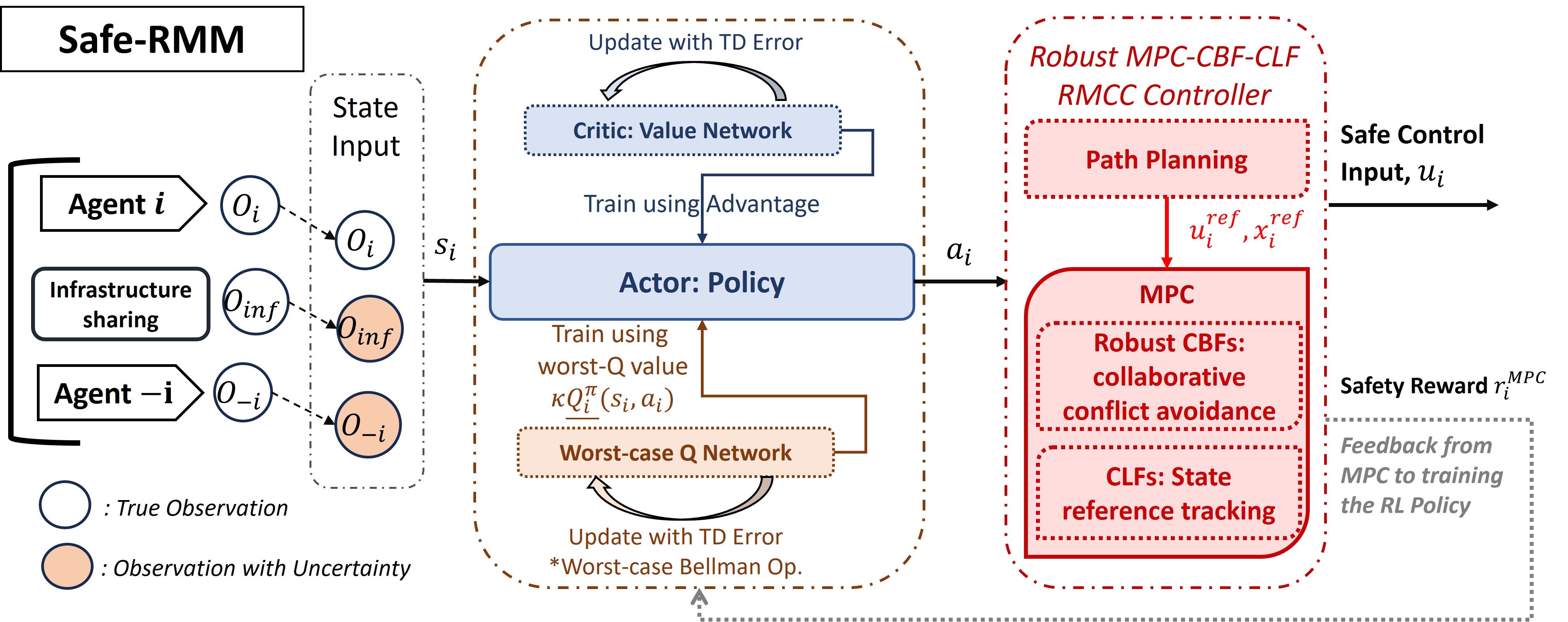}}\vspace{-6pt}
    \caption{Safe-RMM algorithm. The figure demonstrates an agent's decision pipeline while other agents share the same procedure. During training, both Value network and worst-Q network join the update of actor's policy. During testing, Agent $i$ takes states with uncertainty to its actor and samples the high level action $a_i$, which is subsequently handled by robust MPC controller for path-planning and generating safe control $\boldsymbol{u}_{i}$.}
    \label{fig:WMAPPO_RSS}
    \vspace*{-0.4cm}
\end{figure*}

{We present Safe-RMM -- our hierarchical decision-making framework design in this section. We begin by presenting the design of our Robust MARL algorithm and subsequently present the details of the MPC controller using robust CBFs. Our Robust MARL algorithm augments  MAPPO~\cite{yu2022surprising} such that each PPO agent is equipped with a worst-case Q network~\cite{liang2022efficient}. The worst-case Q estimates the potential impact of state perturbations on the policy's action selection and the resulting expected return. By incorporating it into the policy's training objective, we enhance the robustness of the trained policy against state perturbations. Consequently, the proposed algorithm improves both the safety and efficiency of CAVs under state uncertainties. We present our proposed safe MARL algorithm, Safe-RMM, in Algorithm~\ref{alg:SafeRMM}.} 

\setlength{\textfloatsep}{0pt}
\begin{algorithm}[!t]
\caption{Safe-RMM} \label{alg:SafeRMM}
\SetAlgoLined
\textit{Initialize} policy, PPO critic and Worst-case Q networks $\boldtheta_i^0, \boldphi_i^0, \boldomega_i^0$; worst-Q weight $\boldlambda_i^0 = 0$ \;
\For {each episode $E$}{
    \textit{Initialize} $s=\prod_i s_{i}\in \calS$, $\calA =  \prod_i \calA_i$\\
    \textit{Initialize} centralized memory $M=\emptyset$ \\
    $\textbf{\textit{Rollout}}(s, \calA)$: \For{each step, agent $i$}{
        Choose $a_i \in \calA_i$ based on $\epsilon$-greedy, $a= \prod a_i$\;
        Computes safe control and safety reward $\boldsymbol{u}_i^{\textit{safe}}, p_i^{\textit{MPC}} = \textit{MPC}(s_i, a_i)$ (~\ref{sec:Robust_MPC}) \;
        Execute $\boldsymbol{u}_i^{\textit{safe}}$, observe next state $s'=\prod_i s'_{i}$ \; 
        Compute rewards $r=\{r_i\}$ according to (\ref{reward_func})
        Store $(s_i, a_i, r_i, s'_i),\forall i$ in $M$\;
        $s \leftarrow s'$\;
    }
    \textbf{\textit{Training}}: \For {each agent $i$}{
        Compute PPO critic loss gradient $\grad_{\boldphi_i}(\calL^{V}_i)$~\cite{schulman2017proximal}\; 
        Compute worst-Q critic loss gradient $\grad_{\boldomega_i}(\calL^{\wq}_i)$~\cite{liang2022efficient}\; 
        Update $\scriptstyle \boldphi_i^{E}\leftarrow \boldphi_i^{E+1}$,$\scriptstyle\boldomega_i^{E} \leftarrow \boldomega_i^{E+1}$\;
        Compute policy loss gradients $\grad_{\boldtheta_i}(\calL_i(\boldtheta_i))$ (\ref{loss_func})\; 
        Update $\boldtheta_i^{E} \leftarrow \boldtheta_i^{E+1}$, $\boldlambda_i^{E} \leftarrow \boldlambda_i^{E+1}$\;
    }
}
\end{algorithm}

\subsection{Robust MAPPO}
\label{sec:SRMAPPO}
The proposed Robust MARL algorithm (Alg~\ref{alg:SafeRMM}; Fig.~\ref{fig:WMAPPO_RSS}) uses centralized training and decentralized execution. We are inspired by the Worst-case-aware Robust RL framework~\cite{liang2022efficient} and designed the robust MAPPO. Each robust PPO agent maintains a policy network $\iactor{i}$ (``actor'') with parameter $\boldtheta_i$, a value network (``critic'') $V(s)$ with parameter $\boldphi_i$ and the second critic $\iwq{i}(s_i,a_i)$ network approximating \textit{the worst-case action values} with parameter $\boldomega_i$. In order to learn a safe cooperative policy for the CAVs, the MARL interacts with the MPC controller (Sec.~\ref{sec:Robust_MPC}) during the \textbf{\textit{Rollout}} process. As the algorithm starts, agent $i$'s policy takes initial state $s_i$ and samples an action $a_i$; the MPC-controller with robust CBFs given $a_i$ computes a safe control $\boldsymbol{u}_i$ for the vehicle to execute. The agent receives $r_i$~\eqref{reward_func} and all agents synchronously move to the next time-step by observing the new state $s'=\prod_i s'_{i}$.

During training, both critics account for updating the policy with loss function (\ref{loss_func}) so that the trained policy can balance the goals between maximizing expectations of advantage $\hat{A}_i$ and worst-case return $\underline{Q}^{(\boldomega_{i})}$. Through value-based state regularization $\calL^{reg}_i(\boldtheta_i)$, the policy is trained to be robust at crucial ``vulnerable'' states around which uncertainties are more likely to affect the policy~\cite{zhang2020robust, yu2022surprising}.
\begin{align}
\calL_i(\boldtheta_i) &= \scriptstyle\frac{1}{N} \sum_{t}^{N}\min (\rho_{\boldtheta_i}, clip(\rho_{\boldtheta_i}))\textstyle(\hat{A}^{t}_i + \kappa\underline{Q}^{(\boldomega_{i})t})\scriptstyle + \calL^{reg}_i(\boldtheta_i)
\label{loss_func}
\end{align}
\vspace*{-0.6cm}

\subsection{Robust CBF-based Model Predictive Control}
\label{sec:Robust_MPC}

We adopt receding horizon control to implement the low-level controller for every agent $i$ in the road network. The low-level controller maps the high-level plans/actions $a_i \sim \pi_{\boldsymbol{\theta}_i}$ into primitive actions/control inputs for agent $i$. Firstly, a path planning function $z:\mathcal{A} \rightarrow \mathcal{X}_i \times \mathcal{U}_i$ is used to map the high-level plans/actions into state and action references, i.e. $(\boldsymbol{x}_{i}^{ref}, \boldsymbol{u}_{i}^{ref}) = z(\boldsymbol{x}_i, a_i)$, where $\mathcal{X}_i \subset \mathcal{S}_i$ is the state space and $\mathcal{U}_i$ is the input space of primitive actions of CAV $i$ respectively. Subsequently this information is fed to the MPC controller. To prevent collision between agents, safety constraints are incorporated to the controller using CBFs. 

We consider that the dynamics for each vehicle is affine in terms of its control input $\bm{u}$ as follows:
\begin{equation}    
\label{eq_nonlinear}
\boldsymbol{\dot{x}}=f(\boldsymbol{x})+g(\boldsymbol{x})\boldsymbol{u} \nonumber 
\end{equation}
\begin{equation} \label{VehicleDynamics}
\underbrace{\left[\begin{array}{c}
\dot{x} \\
\dot{y} \\
\dot{\psi} \\
\dot{v}
\end{array}\right]}_{\dot{\boldsymbol{x}}(t)}=\underbrace{\left[\begin{array}{c}
v \cos \psi \\
v \sin \psi \\
0 \\
0
\end{array}\right]}_{f\left(\boldsymbol{x}(t)\right)}+\underbrace{\left[\begin{array}{cc}
0 & 0 \\
0 & 0 \\
0 & v / (l_f+l_r) \\
1 & 0
\end{array}\right]}_{g\left(\boldsymbol{x}(t)\right)} \underbrace{\left[\begin{array}{c}
u \\
\phi
\end{array}\right]}_{\boldsymbol{u}(t)},
\end{equation}
where $f$ and $g$ are locally Lipschitz, $\boldsymbol{x}\in \mathcal{X} \subset \mathcal{S}$ denotes the state vector and $\bm{u} \in \mathcal{U} = [\boldsymbol{u}_{min}, \boldsymbol{u}_{max}]$ denotes the control input. In the above equations $x(t), y(t), \psi(t), v(t)$ represent the current longitudinal position, lateral position, heading angle, and speed, respectively. $u(t)$ and $\phi(t)$ are the acceleration and steering angle of vehicle at time $t$, respectively, $g\left(x(t)\right)=$ $\left[g_u\left(\boldsymbol{x}(t)\right), g_\phi\left(\boldsymbol{x}(t)\right)\right]$. 

\begin{figure}
    \centering
    \includegraphics[width=0.8\linewidth]{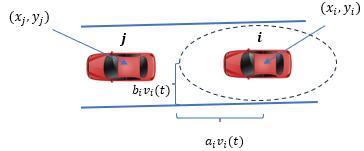}
    \vspace*{-10pt} 
    \caption{Illustration of the ellipsoidal safety set.}
    \vspace*{4pt} 
    \label{fig:safety_boundary}  
\end{figure}

We incorporate safety with respect to other vehicles (primarily unconnected vehicles) as follows. Let CAV $i$ needs to stay safe to a vehicle $j \in \{o_i, o_{\calN_i}, o_{\calN_i^{UV}}\}$ in its vicinity. To achieve that we enforce a constraint on vehicles $i$ by defining a speed dependent ellipsoidal safe region $b_{1}\left(\boldsymbol{x}_i, \boldsymbol{x}_j\right)$ as follows:
\begin{equation}\label{safety}
    b\left(\boldsymbol{x}_i, \boldsymbol{x}_{j}\right):=\frac{\left(x_i(t)-x_{j}(t)\right)^2}{\left(a_i v_i(t)\right)^2}+\frac{\left(y_i(t)-y_{j}(t)\right)^2}{\left(b_i v_i(t)\right)^2}-1 \geq 0,
\end{equation}
where $a_i$, $b_i$ are weights adjusting
the length of the major and minor axes of the ellipse as illustrated in Fig. \ref{fig:safety_boundary}. We enforce safety constraints on any CAV $i$ for vehicles in $\{o_i, o_{\calN_i}, o_{\calN_i^{UV}}\}$ in one of three scenarios which are: i. immediately preceding in the same lane, ii. located in the lane the ego is changing to and iii. arriving from another lane that merges to the lane of the ego vehicle. 

Given a continuously differentiable function $b:\mathbb{R}^n\rightarrow \mathbb{R}$ and the safe set defined as $C:=\{\boldsymbol{x} \in \mathbb{R}^n:b(\boldsymbol{x})\geq 0\}$, $b(\boldsymbol{x})$ is a candidate control barrier function (CBF) for the system (\ref{eq_nonlinear}) if there exists a class $\mathcal{K}$ function $\alpha$ and $\boldsymbol{u}$ such that
\begin{equation}
    L_{f} b(\boldsymbol{x})+L_{g} b(\boldsymbol{x}) \boldsymbol{u}+\alpha(b(\boldsymbol{x})) \geq 0,
    \label{cbf_condition}
\end{equation}
for all $\boldsymbol{x} \in C$, where $L_{f}, L_{g}$ denote the Lie derivatives along $f$ and $g$, respectively. Additionally, we use CLFs to incorporate state references to the controller. A continuously differentiable function $V:\mathbb{R}^n \rightarrow \mathbb{R}$ is a globally and exponentially stabilizing CLF for \eqref{eq_nonlinear} if there exists constants  $c_i \in \mathbb{R}_{>0}$, $i=1,2$, such that $c_1 ||\bm{x}||^2 \leq V(\bm{x}) \leq c_2 ||\bm{x}||^2$, $\boldsymbol{u} \in \mathcal{U}$, and the following inequality holds
\begin{equation} \label{CLF}
L_fV(\boldsymbol{x})+L_gV(\boldsymbol{x})u+ \eta(\boldsymbol{x})\leq e,
\end{equation}
where $e$ makes this a soft constraint.

The uncertain state measurements stemming from process noise and measurement noise denoted by $\hat{\bm{x}}(t)$ is expressed as follows:
\begin{equation}
\label{state_uncertainty}
 \hat{\bm{x}}(t) = \bm{x}(t) + \bm{w}(t)   
\end{equation}
where $\bm{w}(t)$ is bounded noise such that $\|\bm{w}(t)\|_\infty \le \epsilon$. In the presence of noise, the robust CBF constraint becomes the following which has been shown to make the safety set $C$  forward invariant in \cite{ahmad}.
\begin{align} \label{robust_trust_cbf_condition}
&\min_{\scriptstyle {\{\bm{w}(t): \|\bm{w}(t)\|_{\infty} \leq \epsilon\}}} [L_fb(\scriptstyle \boldsymbol{\hat{x}}(t) - \bm{w}(t) \textstyle)]+ L_gb(\scriptstyle\boldsymbol{\hat{x}}(t)     - \bm{w}(t)\textstyle)\boldsymbol{u}(t) \nonumber \\
&+ \alpha(b(\boldsymbol{\hat{x}}(t) - \bm{w}(t)))] \geq 0 
\end{align}

Finally, the MPC with robust CBF and CLF control problem can be expressed as follows:
\begin{align}
&\displaystyle\min_{\boldsymbol{x}_{0:N|k} \atop \boldsymbol{u}_{0:N-1|k}} \sum_{h=0}^{N-1} \bigg((\boldsymbol{x}_{h|k} - \bm{x}_{h|k}^{ref})^T  A_x (\boldsymbol{x}_{h|k} - \bm{x}_{h|k}^{ref}) \nonumber\\
&+ (\boldsymbol{u}_{h|k} - \bm{u}_{h|k}^{ref})^T A_u  (\boldsymbol{u}_{h|k} - \bm{u}_{h|k}^{ref})
+ \bm{B}^T \begin{bmatrix}
\boldsymbol{x}_{h|k}\\ 
\boldsymbol{u}_{h|k}
\end{bmatrix} + \nonumber\\
& \bm{\delta}^T\bm{e}_{h|k}^2  \bigg) + V_N(\boldsymbol{x}_{N|k}) \nonumber
\label{MPC-CBF}
\end{align}
\begin{align}
    \textnormal{subject to} &  \ \   \boldsymbol{x}_{h+1|k}=f(\boldsymbol{x}_{h|k},\boldsymbol{u}_{h|k}),    \ \ \ \ \ h=0,\dots,N-1, \nonumber \\ 
    &\eqref{cbf_condition}, \eqref{CLF}, \ \ \ \ \ \ \ \ \ \ \ \ \ \ \ \ \ \ \ \ \ \ \ \ \ \ \ h=0,\ldots,N-1, \nonumber \\ 
    & \boldsymbol{u}_{h|k} \in \mathcal{U}, \boldsymbol{x}_{h|k} \in \mathcal{X}, x_N \in \mathcal{X}_f \ h=0,\ldots,N-1 \nonumber
\end{align}
where $A_x \in \mathbb{R}^{n \times n}$, $A_u \in \mathbb{R}^{q \times q}$ are the matrix of weights, $\bm{B} \in \mathbb{R}^{n+q}$ is vector of weights, and $\bm{\delta}$ is the vector of weights of the penalty terms associated with the relaxation parameters of the CLF constraints.

\section{Experiments and Evaluations}
\label{sec:experiment}
\begin{figure}[!t]
\centering
\scalebox{0.85}
{
    \subfloat[]
    {\centering\label{fig:cross_scene}\includegraphics[height=2.9cm,width=8.8cm]{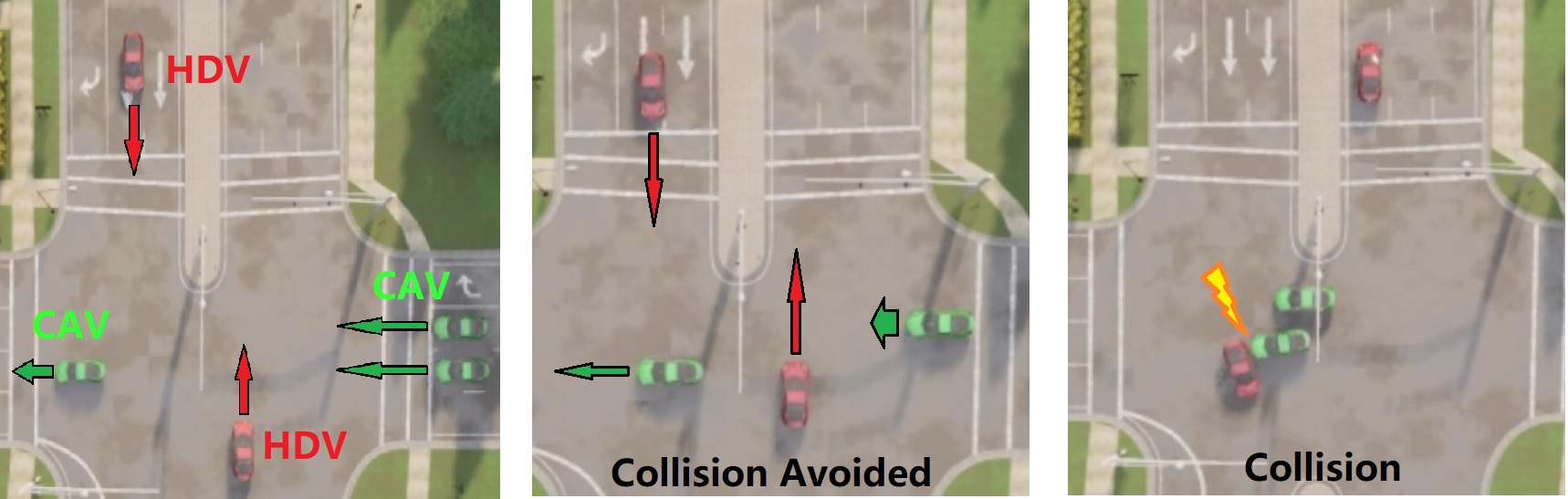}}
}
\scalebox{0.85}
{
    \subfloat[]{\centering\label{fig:hwy_scene}\includegraphics[height=2.4cm,width=8.8cm]{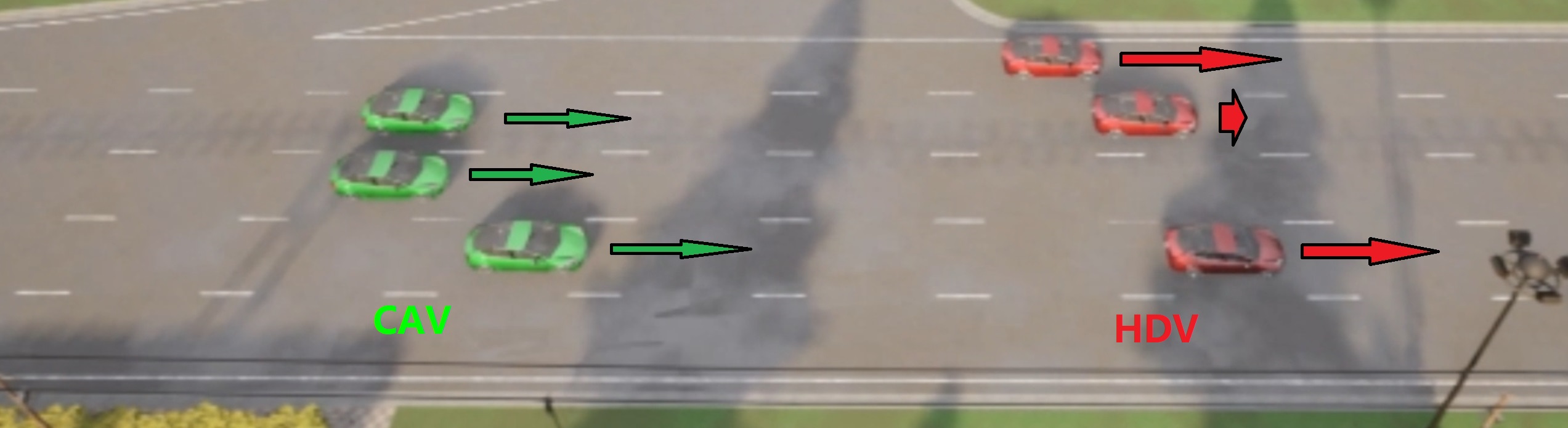}}
}
\caption{\textit{Intersection} (a) and \textit{Highway} (b) scenarios for testing. (\ref{fig:cross_scene}) 3 CAVs and 2 HDVs participate (left), CAVs could either dodge (middle) or collide with HDVs (right); (\ref{fig:hwy_scene}) 3 CAVs and 3 HDVs participate in a multi-lane Highway scenario, with one HDV suddenly brakes.}
\label{fig:intersection_cases}
\end{figure}

We conduct our experiment in the CARLA Simulator environment~\cite{Dosovitskiy17}, where each vehicle is configured with onboard GPS and IMU sensors and a collision sensor that detects the collision with other objects. We show two challenging scenarios in daily driving, respectively, at \textit{Intersection} (Fig.~\ref{fig:cross_scene}) and on \textit{Highway} (Fig.~\ref{fig:hwy_scene}), where we spawn multiple CAVs and some HDVs randomly. We adopt three types of state uncertainties exclusively for testing: A random error ${e^{\text {rand}}\scriptstyle \sim \textit{U}(-3,3)}$ (\textit{U}: uniform distribution); \textit{error\_over\_time}: $\mathtt{ERR}^{T}$ and \textit{error\_target\_vehicles}: $\mathtt{ERR}^{\calV}$ that impose perturbation at consistent values to CAVs' states to affect their behavior patterns:
\begin{align*}
    \mathtt{ERR}^{T} &= \{(e^t, e^t)| e^t{\scriptstyle\sim\textit{U}(e^0-\frac{1}{2},e^0+\frac{1}{2})},\pm e^0 {\scriptstyle\sim \textit{U}(4,5)}, t\in T\} \\
    \mathtt{ERR}^{\calV} &= \{(e^{\nu}, e^{\nu})| e^{\nu}{\scriptstyle\sim\textit{U}(e^0-\frac{1}{2},e^0+\frac{1}{2})},\pm e^0 {\scriptstyle\sim \textit{U}(4,5)}, {\nu}\in \calV\}
\end{align*}
The former $\mathtt{ERR}^{T}$ has state errors for all cars in duration $T \subset E$, while the latter $\mathtt{ERR}^{\calV}$ has a subset of vehicles ${\calV}$ randomly sampled in each episode and adds uncertainties to how ${{\nu}\in \calV}$ are observed by others throughout the current episode.

\paragraph{Intersection} The Fig.~\ref{fig:Scenarios} and Fig.~\ref{fig:cross_scene} present an snapshot of the \textit{Intersection} scenarios with CAVs (green) passing the intersection and HDVs (red) from opposite sides crossing the box at the same time. HDVs pose critical safety threats as they could either hit or be hit by a CAV from the side when driving fast ($\approx 10 m/s$). CAVs aim to avoid collision and reach the preset destination after passing through the intersection. 

\paragraph{Highway} Figures.~\ref{fig:hwy_scene} illustrates the scenario, where CAVs (green) are spawned behind HDVs (red) on a multi-lane highway. During training and evaluation, HDVs keep in their lanes at random speed from [7-9] $m/s$ except one random HDV simulates a stop and go scenario. CAVs aim to avoid any collision and drive at the speed limit of the road to arrive at the destination. 

\begin{figure}[h!]
    \centering
    \includegraphics[width=0.80\linewidth]{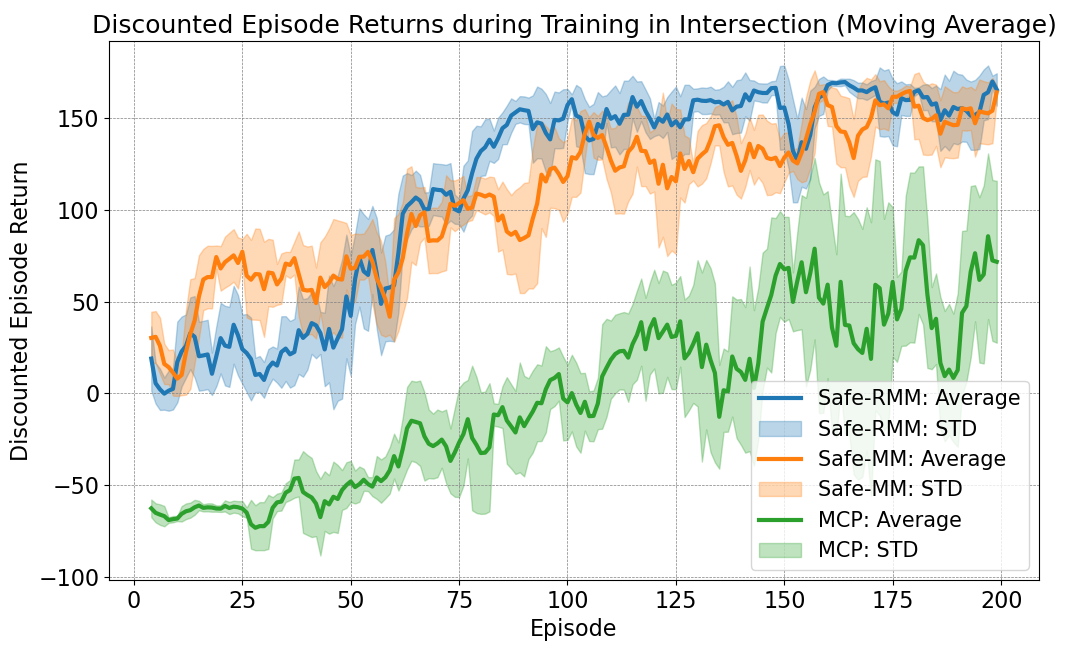}
   \vspace*{-6pt} \caption{Discounted Efficiency Returns during Training in \textit{Intersection}.}  
    \label{fig:train_fig}  
\end{figure}

\begin{figure}[h!]
    \centering
    \includegraphics[width=0.80\linewidth]{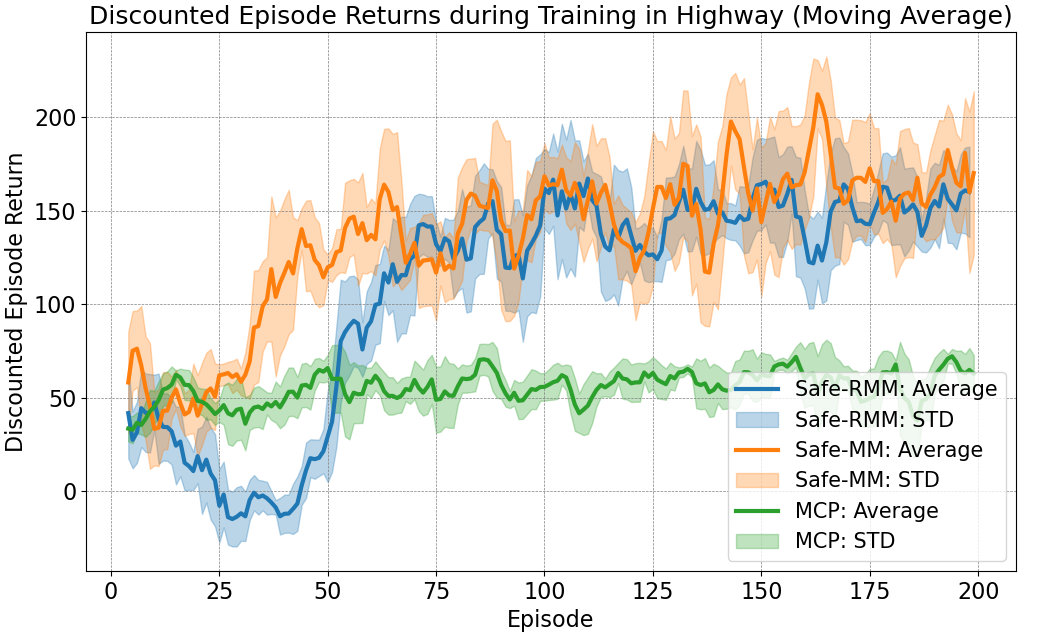}
   \vspace*{-6pt} \caption{Discounted Efficiency Returns during Training in \textit{Highway}.}  
    \label{fig:train_highway_fig}  
\end{figure}

\subsection{Experiment Results}
 In this section, we highlight our method's performances in terms of safety guarantees and robustness against state uncertainty while generalizable to different driving scenarios. We demonstrate through ablation studies that the incorporation of MPC with robust CBF based controller improves the performance of MARL algorithm. Additionally, we also demonstrate that our proposed hierarchical approach with robust MARL also improves the MPC-CBF controller. 
 
 We trained three models: our Safe-RMM method, the ``non-robust'' Safe-MM that also adopts our framework, and the MCP method adopting MARL-PID controller with CBF safety shield~\cite{zhang2023spatial}. Each model is trained on \textit{Intersection} and \textit{Highway} respectively for 200 episodes. In evaluation, aside from the three trained models, we have ``MP'', MARL with PID controller, as an example of learning-based method without safety shielding; and we also implemented the benchmark ``RULE'' adopting a rule-based planner and a robust MPC controller. The ``RULE'' benchmark has been implemented based on the method proposed in \cite{10422265}, which introduces a safety-guaranteed rule for managing vehicle merging on roadways. This rule ensures safe interactions between vehicles arriving from different roads converging at a common point. Methods are evaluated for 50 episodes in both scenarios under four uncertainty configurations: None (uncertainty-free), random error $e^{\text{rand}}$, and two targeting errors $\mathtt{ERR}^{\calV}$ and $\mathtt{ERR}^{T}$. Training results in \textit{Intersection} are shown in Fig.~\ref{fig:train_fig}; evaluations in both scenarios are presented in Table.~\ref{tab:Eval_crossing_highway}. For each entry in both tables, the left integer is the number of collisions happened during evaluation (in 50 episodes); the right number is the agents' mean discounted return considering only the rewards related to velocity and goal-achievement in (\ref{reward_func}). We highlight the top performance across all methods with the least collision numbers and the highest efficiency return.

\begin{table}[!t]
\centering
\renewcommand{\arraystretch}{1.35}
\caption{Evaluation Results in Intersection and Highway}
\begin{tabularx}{0.96\columnwidth}{Xcccc}
\hline
        & \multicolumn{4}{c}{Uncertainty} \\ \cline{2-5}
Method & \textbf{None} & \textbf{$e^{\text{rand}}$} & \textbf{$\mathtt{ERR}^{\calV}$} & \textbf{$\mathtt{ERR}^{T}$} \\ \hline
\multicolumn{5}{c}{\textit{Intersection}} \\ \hline
\textbf{Safe-RMM}$^1$  & \textbf{0}, \textbf{162.9}   & \textbf{0}, \textbf{161.4} & \textbf{0}, \textbf{162.2} & \textbf{0}, \textbf{161.8}  \\ 
\textbf{Safe-MM}$^2$   & \textbf{0}, 157.9  & \textbf{0}, 155.7   & \textbf{0}, 155.9 & \textbf{0}, 155.7  \\ 
\textbf{MCP}$^3$  & 3, 65.7  & 2, 60.6 & \textbf{0}, 66.2   & 2, 67.7                \\ 
\textbf{MP}$^4$   & 33, 148.4	& 41, 149.1	& 36, 145.9	& 30, 139.0 \\ 
\textbf{RULE}$^5$ & 2, 120.9 & 1, 113.9  & 3, 105.5 & 2, 112.3 \\ \hline
\multicolumn{5}{c}{\textit{Highway}} \\ \hline
\textbf{Safe-RMM}   & \textbf{0}, \textbf{162.0} & \textbf{0}, \textbf{169.4} & \textbf{0}, 166.4 & \textbf{0}, 161.8 \\ 
\textbf{Safe-MM}    & \textbf{0}, 161.3 & \textbf{0}, 168.7 & \textbf{0}, \textbf{168.7} & \textbf{0}, \textbf{163.0} \\ 
\textbf{MCP}   & \textbf{0}, 56.8  & 2, 55.8  & 1, 60.7  & 2, 58.4  \\ 
\textbf{MP} & 35, 74.1 & 34, 74.5 & 38, 73.8	& 38, 74.5 \\ \hline
\end{tabularx}
\begin{tablenotes}
\item $^1$Safe-RMM: our method--Safe Robust MARL-MPC; $^2$Safe-MM: Safe MARL-MPC adopting same framework as$^1$ but trained without worst-case Q. Benchmarks: $^3$MCP: MARL-PID with CBF Safety Shield; $^4$MP: MARL-PID without shielding; $^5$RULE: rule-based planner with MPC controller.
\item Each entry in the table above contains (\textit{number of collisions}; \textit{mean of episodes discounted efficiency return}).
\end{tablenotes}
\label{tab:Eval_crossing_highway}
\end{table}

\subsubsection{Top Safety and Efficiency Achieved by the Framework}
Our proposed MARL-MPC framework demonstrates top safety performance as both Safe-RMM and Safe-MM achieve zero collisions across all evaluation scenarios, even when subjected to uncertainties. Additionally, Safe-RMM and Safe-MM rank among the top two in terms of efficiency across all settings. Our proposed approach with robust MPC controller enables MARL to fully realize its potential, allowing its decisions to be executed with accuracy and receiving precise reward feedback. The reciprocation between the MARL and MPC controller enhances policy training and demonstrates that ensuring safety in autonomous vehicles need not compromise efficiency.

In contrast, the MP baseline, adopting MARL without safety shielding, experiences collisions in 60\%-80\% of evaluation episodes as shown in Table~\ref{tab:Eval_crossing_highway}. The result highlights the limited safety-awareness of pure learning-based approaches in complex driving scenarios. The MCP baseline, which incorporates a CBF-based safety shield, significantly reduces collisions to just 3\% of episodes in average. However, this comes at the cost of performance efficiency due to more conservative behaviors. Specifically, MCP in the presence of safety shield shows a 54\% performance drop in the \textit{Intersection} scenario and a 23\% drop in the \textit{Highway} scenario compared to the MP method. These results verify that when applying learning-based algorithm in CAV system, the policy and the controller (or the action actuator) are not independent. Our Safe-RMM having MARL enabled by the accuracy and safety of MPC controller achieves comprehensively best performance while the same MARL algorithm is hindered by the limited capability of PID controller from reaching its optimality.

\subsubsection{Ablation Study with Rule-based Benchmark and Robustness}
We conducted an ablation study by evaluating the rule-based benchmark ``RULE'' in the \textit{Intersection} scenario. As shown in Table~\ref{tab:Eval_crossing_highway}, our Safe-RMM method outperforms ``RULE'' in both safety and efficiency metrics. Compared with other benchmarks, the rule-based method offers a more ``balanced'' performance -- being significantly safer and less ``reckless'' than MP, while achieving comparable safety and much higher efficiency than MCP. However, in a safety-critical scenario like the \textit{Intersection}, 
more precise decision-making is required for CAVs to ensure both safety and speed. The lack of adaptability and learning capability limits rule-based approaches from achieving comparable performance as our approach in these complex scenarios.

From the results of the RULE benchmark in Table~\ref{tab:Eval_crossing_highway}, we observe a 6\%-13\% drop in efficiency when comparing episodes with and without state uncertainties. This supports our earlier assessment in Sec.~\ref{sec:related} that rule-based methods, as high-level planners, generally lack the robustness of learning-based approaches. Furthermore, a comparison between Safe-RMM and its ``non-robust'' counterpart, Safe-MM, shows that Safe-RMM outperforms Safe-MM by approximately 4\% in efficiency across all evaluation settings in \textit{Intersection}. However, this advantage does not extend to the \textit{Highway} scenario. Our analysis of both scenarios suggests that the uncertainties of HDVs come not only from tested errors, but also from the initial randomization of HDVs' location and speed. In \textit{Intersection}, even slight variations in vehicle states may require completely different optimal strategies by the CAVs -- either seizing the opportunity to pass ahead of the HDVs, or yielding for safety at the cost of efficiency. Safe-RMM demonstrates greater robustness against these uncertainties and can effectively manage these ``critical'' moments, optimizing for higher expected returns with a worst-case consideration. However, in less safety-critical scenarios like \textit{Highway}, the worst-case awareness of Safe-RMM can result in sub-optimality. In these cases, the algorithm may avoid taking greedier actions and have lower efficiency, even when such actions carry little risk.

\section{Conclusion}
\label{sec:conclusion}
In this work, we study the safe and robust planning and control problem for connected autonomous vehicles in common driving scenarios under state uncertainties. We propose the Safe-RMM algorithm for coordinated CAVs and further validate through experiments the effectiveness of our method. MARL can enhance the performance ceiling of MPC controller in safety-critical scenarios and robust MPC controller can safely and accurately execute the actions from RL and reciprocally contributes to better trained policies. The method achieves top safety and efficiency performances in evaluation, and maintain robustness against tested perturbations. Future work could consider optimizing the control policy in a mixed traffic scenario with both RL and rule-based intelligent agents.


\bibliographystyle{IEEEtran}
{\small \bibliography{sec_08_Ref}}


\end{document}